\documentclass[3p,times]{elsarticle}
\usepackage{amssymb}
\usepackage{pifont}
\usepackage{geometry}
\usepackage{txfonts}
\usepackage[table,x11names]{xcolor}
\usepackage[hidelinks]{hyperref}
\usepackage{textcomp}
\usepackage{amsthm}
\usepackage{amsmath}
\usepackage[T1]{fontenc}
\usepackage{float}
\usepackage{url}
\usepackage{times,amsmath,epsfig,cite}
\usepackage[linesnumbered,ruled]{algorithm2e}
\usepackage{graphics,graphicx}
\usepackage{multirow}
\usepackage{setspace, caption}
\journal{Neurocomputing}
\begin{document}
\begin{frontmatter}
\title{Segmentation of Glioma Tumors in Brain Using Deep Convolutional Neural Network}
\author[1]{Saddam Hussain}
\author[1]{Syed Muhammad Anwar  \corref{cor1}}
\cortext[cor1]{Corresponding author}
\ead{s.anwar@uettaxila.edu.pk}
\author[2]{Muhammad Majid}
\address[1]{Department of Software Engineering, University of Engineering \& Technology, Taxila, 47050 Pakistan}
\address[2]{Department of Computer Engineering, University of Engineering \& Technology, Taxila, 47050 Pakistan}

\begin{abstract}

Detection of brain tumor using a segmentation based approach is critical in cases, where survival of a subject depends on an accurate and timely clinical diagnosis. Gliomas are the most commonly found tumors having irregular shape and ambiguous boundaries, making them one of the hardest tumors to detect. The automation of brain tumor segmentation remains a challenging problem mainly due to significant variations in its structure. An automated brain tumor segmentation algorithm using deep convolutional neural network (DCNN) is presented in this paper. A patch based approach along with an inception module is used for training the deep network by extracting two co-centric patches of different sizes from the input images. Recent developments in deep neural networks such as drop-out, batch normalization, non-linear activation and inception module are used to build a new ILinear nexus architecture. The module overcomes the over-fitting problem arising due to scarcity of data using drop-out regularizer. Images are normalized and bias field corrected in the pre-processing step and then extracted patches are passed through a DCNN, which assigns an output label to the central pixel of each patch. Morphological operators are used for post-processing to remove small false positives around the edges. A two-phase weighted training method is introduced and evaluated using BRATS $2013$ and BRATS $2015$ datasets, where it improves the performance parameters of state-of-the-art techniques under similar settings.
\end{abstract}
\end{frontmatter}

\section[INTRODUCTION]{Introduction}
\label{sec:introduction}
In the age of machines, where more and more tasks are being automated, the automation of image segmentation is of substantial importance. This is also significant in the field of medicine, due to the sensitivity of underlying information. Segmentation of lesions in medical imaging provide invaluable information for lesion analysis, observing subject's condition and devising a treatment strategy. Brain tumor is an abnormality in brain tissues leading to severe damage to the nervous system, which in extreme cases can lead to death. Gliomas are the most common and threatening brain tumors with the highest reported mortality rate due to their quick progression \citep{1}. Gliomas tumors are infiltrative in nature and mostly escalate near the white matter fibre, but they can spread to any part of the brain making them very difficult to detect. Gliomas tumors are generally divided into four grades by the world health organization (WHO) \citep{2}. Grade one and grade two tumors refer to the low grade gliomas (LGG), whereas grade three and grade four are known as the high grade gliomas (HGG), which are severe tumors with a life expectancy of about two years \citep{3}. Grade four tumors are additionally called glioblastoma multiforme (GBM) \citep{4} and have an average life expectancy of around one year \citep{5}. GBM and the encompassing edema can lead to a major impact, devouring healthy tissues of the brain. High grade gliomas show conspicuous micro-vascular multiplications and territories of high vascular thickness. Treatment alternatives for gliomas incorporate surgery, radiation treatment, and chemotherapy \citep{6,7}.

Magnetic resonance imaging (MRI) is a commonly used imaging technique for detection and analysis of brain tumors. MRI is a non-intrusive system, which can be utilized alongside other imaging modalities, such as computed tomography (CT), positron emission tomography (PET) and magnetic resonance spectroscopy (MRS) to give an accurate data about the tumor structure \citep{1,8}. However, use of these systems alongside MRI is expensive and in some cases can be invasive such as PET. Therefore, different MRI modalities that are non-invasive and image both structure and functions are mostly used for brain imaging. MRI machines themselves come with different configurations and produce images with varying intensities. This makes tumor detection a difficult task when different MRI configurations (such as 1.5, 3 or 7 Tesla) are used. These configurations have different intensity values across voxels, which result in masking the tumor regions \citep{6}. MRI can be normalized to harmonize tissue contrast, making it an adaptable and widely used imaging technique for visualizing regions of interest in the human brain. MRI modalities are combined to produce multi-modal images giving more information about irregular shaped tumors, which are difficult to localize with a single modality. These modalities include T1-weighted MRI (T1), T1-weighted MRI with contrast improvement (T1c), T2-weighted MRI (T2) and T2-weighted MRI with fluid attenuated inversion recovery (T2-Flair) \citep{9}. This multi-modal data contains information that can be used for tumor segmentation with significant improvement in performance.

The human brain is usually segmented into three regions i.e., white matter (WM), grey matter (GM) and cerebrospinal fluid (CSF) \citep{10}. Tumor regions normally reside near the white matter fibre and have fuzzy boundaries, making it a challenging task to segment them accurately. Different tumor regions include necrotic center, active tumor region, and edema, which is the surrounding area swelled by the effects of tumor. A correctly segmented tumor region is significant in medical diagnosis and treatment planning, hence it has drawn huge focus in the field of medical image analysis \citep{11, 12}. Manual segmentation of glioma tumors across the MRI data, while dealing with an increasing number of MRI scans is a near to impossible task. Therefore, many algorithms have been devised for automatic and semi-automatic segmentation of tumors and intra-tumor structures. Historically, standardized datasets have not been available to compare the performance of such systems. Recently, the medical image computing and computer assisted intervention society (MICCAI) has started a multi-modal brain tumor segmentation challenge (BRATS), held annually, providing a standard dataset that is now being used as a benchmark for the evaluation of automated brain tumor segmentation task \citep{13}.

Detecting tumor in MR images is a difficult problem, as two pixels can have very similar features but different output labels. Structured approaches are available, such as conditional random field (CRF) that deal with multiple predictions by taking context into account. These approaches are computationally expensive, both in terms of time and resources. The segmentation of cerebral images is generally categorized into voting strategies, atlas, and machine learning based grouping techniques \citep{17}. Recently, machine learning algorithms have gained popularity through their efficient and accurate predictions in segmentation tasks. It is a common practice to use hand crafted features that are passed to a classifier to predict an output class in machine learning based methods. Most machine learning techniques fall in the category of probabilistic methods, which compute the probabilities of possible output classes based on given input data. The class with highest predicted probability is assigned as a label to the input sample. Usually, probabilistic methods are used after obtaining anatomical models by mapping brain atlases on $3D$ MR images \citep{19}. Machine learning methods are further divided into discriminative and generative approaches \citep{18}. 

Generative models rely heavily on prior historical or technical data that come from surroundings. In the MR image segmentation task, this calls for a need to take large portions of image into account, since tissues have irregular shape and tumors have varying structures. These generative models are optimized to give good performance using sparse data maximum-likelihood (ML) estimation \citep{20, 21}. On the contrary, discriminative approaches use very little prior data and rely mostly on large number of low level image features. Image processing techniques based on raw pixels values \citep{22, 23}, global and local histograms \citep{24, 25}, texture and alignment based features etc., fall in the category of discriminative approaches. Techniques, such as random forest (RF) \citep{26}, support vector machine (SVM) \citep{27}, fuzzy C-means (FCM) \citep{29}, and decision forests \citep{30} have been used for brain tumor segmentation. These methods have limited capacity and in most cases do not provide results, which can be used for clinical purposes. SVM is an effective algorithm for binary class classification, whereas in most cases tumor is classified into multiple classes. FCM is a clustering based technique and do not rely on the assigned labels. Decision trees are based on binary decision at every node, thus they are affected by over-fitting. Random forest based techniques reduce the over-fitting problem, but they are useful for unsupervised learning. 

Deep learning based techniques are a good prospect for image segmentation, especially convolutional neural networks (CNNs) are tailor made for pattern recognition tasks. Neural networks learn features directly from the underlying data in a hierarchical fashion, in contrast to the statistical techniques such as SVM, which rely on hand crafted features \citep{36}. Deep neural networks have been successfully applied for medical image analysis tasks such as image segmentation \citep{33, 34, Valverde2017159, }, retrieval \citep{Qayyum2017} and tissue classification \citep{35}. Recently, pixel based classification is gaining popularity as it is expected to give absolute accuracy if every pixel is properly classified. 

Markov random field (MRF) combined with sparse representation has been utilized for dealing with variability problems in MR images \citep{37}. Maximum a posterior (MAP) probability has been calculated using likelihood and MRF, which is then used to predict the tumor classes using grow-cut. An ensemble of 2D CNNs alongside growcut has been used to estimate the tumor area \citep{38}. In \citep{39}, two classifiers have been used to compute probabilities of tumor and background classes. The first classifier named global classifier needs to be trained only once while the second classifier named custom classifier is tuned with respect to every test image. In \citep{40}, patches are extracted from every voxel in a $3D$ image and a CNN is trained over these patches. A feature array is extracted from the last fully connected layer of the network and is further used to train RF based classifier to predict output labels for tumor as well as healthy pixels. A structured approach to segment MR image pixels into output classes is presented in \citep{41}. The segmentation task is divided into a pair of sub-tasks, where clusters are formed for output predictions and a CNN is trained to classify pixels into output clusters. These methods need an iterative training procedure, requiring sufficient number of training images and are effected by the data imbalance problem in the pixel based brain tumor segmentation task. Also, linear CNN architectures become computationally expensive as the number of convolution layers are increased. The selection of optimal hyper-parameters is also a challenging task that needs to be addressed carefully.
 
In this study, multiple deep learning based architectures for brain tumor segmentation are presented. A patch based approach is used to classify individual pixels in an MR image exploiting advances in the field of neural networks, such as drop-out \citep{14}, batch normalization \citep{15} and  max-out \citep{16}. The nexus architectures are proposed in this work that exploit both the contextual and local features, while predicting the output label and still keep the computational cost within an acceptable range. Larger kernels are used in the first part of the nexus networks to exploit context, while smaller kernels are used in the later part of the networks to exploit local features. The experiments are performed on BRATS $2013$ and BRATS $2015$ datasets. The proposed method achieves improved results on all major evaluation parameters when compared with state-of-the-art techniques. The key contributions of this study are,

\begin{itemize}
\item A fully automated method for brain tumor segmentation is proposed, which uses recent advances in convolutional neural networks, such as drop-out, max-out, batch normalization, inception module and proposes a two phase weighted training method to deal with data imbalance problem.
\item The proposed method considers both the local and contextual information, while predicting output labels and is efficient compared to the most popular structured approaches.
\item The proposed method takes $5$ to $10$ minutes to segment a brain, achieving improvement of an order of magnitude compared to most state-of-the-art techniques.  

\end{itemize}

The remainder of this paper presents, the proposed method in \autoref{sec:method}, experimental set up in \autoref{sec:experiment}, experimental results and discussion in \autoref{sec:results}, followed by conclusion in \autoref{sec:conclusion}.
  
\section[PROPOSED METHODOLOGY]{Proposed Methodology}
\label{sec:method}

\begin{figure*}[!t]
  \centering
  \includegraphics[width=4.5in, height=1.2in]{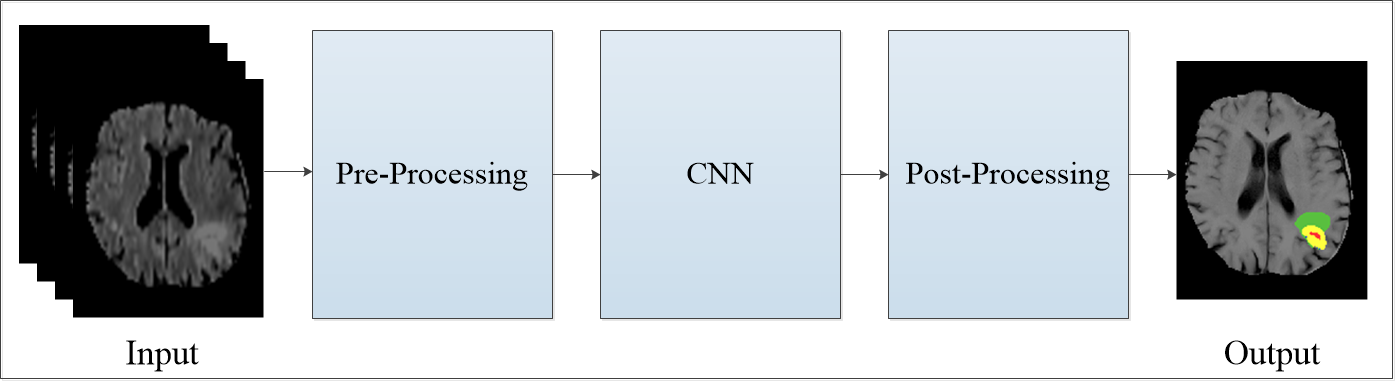}
  \caption{Block diagram of the proposed methodology.}\label{fig:fig1}
\end{figure*}

The proposed methodology elaborates on the effectiveness of convolutional neural networks in brain tumor segmentation task. The proposed methodology consists of three steps i.e., pre-processing, CNN, and post-processing as shown in \autoref{fig:fig1}. The input images are pre-processed and divided into patches, which are then passed through a convolutional neural network to predict the output labels for individual patches. The details of each step are as follows.

\subsection{Pre-processing}
\begin{figure}[t]
\centering
\includegraphics[width=2.2in, height=1.7in]{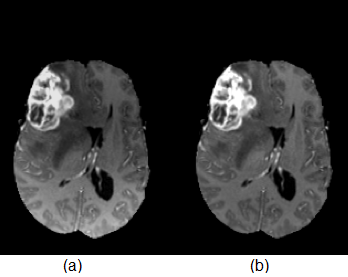}
\caption{MRI scan (a) before (b) after N4ITK bias field correction.}
\label{fig:fig2}
\end{figure}

Images acquired from different MRI modalities are affected by artefacts, such as motion and field inhomogeneity \citep{42}. These artefacts cause false intensity levels, which leads to the emergence of false positives in the predicted output. Bias field correction techniques are used to deal with artefacts in MRI. The non-parametric, non-uniform intensity normalization (N3) algorithm is a widely used method for intensity normalization and artefact removal in MR images \citep{13}. An improved version of N3 known as N4ITK bias field correction  \citep{43} is used in this study to remove unwanted artefacts from MR images. The 3D slicer tool-kit version $4.6.2$ is used to apply bias field correction, which is an open source software that provides tools to visualize, process, and give important information regarding 3D medical images \citep{44, 45}. The effects of intensity bias in a MR image and the result of applying bias field correction is shown in \autoref{fig:fig2}. Higher intensity values are observed in the first scan near bottom left corner, which can lead to false positives in automated segmentation. The second scan shows better contrast near the edges by removing bias using N4ITK bias field correction.

It has been observed that the intensity values across MRI slices vary greatly, therefore normalization is applied in addition to the bias field correction to bring mean intensity value and variance close to zero and one, respectively. The top and bottom one percent intensity values are also removed in addition to the normalization process, which brings the intensity values within a coherent range across all images and facilitate learning in the training phase. A normalized slice $x_{n}$ is generated as follows,
\begin{equation}
  x_{n} = \frac{x - \mu}{\sigma} , 
  \end{equation}
where $x$ represents the original slice, $\mu$ and $\sigma$ are the mean and standard deviation of $x$ respectively. 

For an input concatenated architecture, two different sized patches are extracted from the slices. A bigger patch of size $M \times M$ and a smaller patch of size $m \times m$ co-centric with the bigger patch is used. The patches are also normalized with respect to mean and variance, such that mean approaches zero and the variance approaches one. 

\subsection{Convolutional Neural Networks}
\begin{figure*}[!t]
  \centering
  \includegraphics[width=5.7in, height=1.9in]{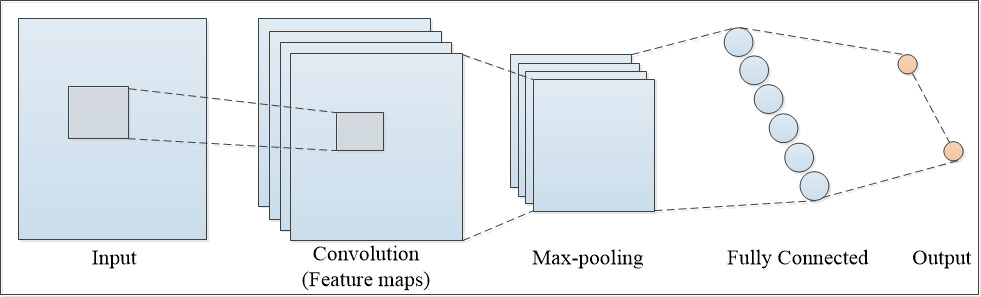}  
  \caption{A simple convolutional neural network architecture showing a feed forward pass with one convolution, max-pooling and fully connected layer.}\label{fig:fig3}
\end{figure*}

A CNN consists of multiple layers such as pooling, convolution, dense, drop-out and so on. The convolution layers are the main building block of a CNN. These layers are stacked over one another in a hierarchical fashion forming the feature maps. Each convolution layer takes feature maps as an input from its preceding layer, except the first convolution layer, which is directly connected to the input space. A convolution layer generates a number of feature maps as output. The CNNs have the ability to learn complex features by forming a hierarchy of feature maps, which makes them very useful and efficient tool for learning patterns in image recognition tasks. A simple configuration of a convolutional neural network is shown in \autoref{fig:fig3}. The convolution layer kernels are convolved over the input sample computing multiple feature maps. A feature is detected from input sample, represented by a small box in the feature maps. These maps are passed to the max-pooling layer, which retains the relevant features and discards the rest. The features from the max-pooling layer are converted into one dimensional feature vector in the fully connected layer, which are then used to compute the output probabilities. A feature map in the network corresponds to a grouping of hidden units called neurons, which are controlled with an activation function. The activations are influenced by neighbouring voxels. Area that affects the activation is called the neuronal receptive field, which increases in each subsequent convolution layer. Each point in a convolutional map is associated with the preceding layers via weights on the connections.

In these layers, non-linear feature extractors, also known as kernels are used to extract features from different input planes. These kernels are convolved with the input planes in a sliding window fashion. The responses of the convolution layer kernels are arranged in a topological manner in the feature maps. The proposed model processes patches belonging to four MRI modalities namely, T1, T1c, T2 and T2-Flair to predict the output label for each pixel in the corresponding patch, performing segmentation of the entire brain. These MRI modalities are provided as an input to first layer of the proposed deep convolutional neural network. The subsequent layers take feature maps generated by preceding layer as input. These MRI modalities behave in a manner, similar to the red, green and blue planes of a color image. A feature map $O_{a}$ is obtained as, 
\begin{equation}
  O_{a}= b_{a} + \sum_{r}F_{ar} * I_{r} ,  
\end{equation}

  where $F_{ar}$ is the convolution kernel, $I_{r}$ is the input plane, $b_{a}$ represents the bias term and the convolution operation is represented by $*$.

The weights on connections as well as kernels are learned through a method called back-propagation \citep{48}. In back-propagation, input images are passed through the network i.e., feed forward pass and predictions are compared to the output labels. The weights are updated starting from the last layer and moving back towards the input. A CNN produces translation invariant feature maps, therefore same feature is detected for the entirety of data by a single kernel in a convolution layer. These networks learn features directly from the image as opposed to most classifiers, which take a feature vector as input. A CNN kernel can be designed having different dimensions such as $3\times3$, $5\times5$ and $7\times7$ etc. The varying kernel shapes take in to account the underlying contextual information. The kernels in a convolution layer have been noted to resemble edge detectors, learning different features according to the properties of training data. 

The CNN architecture deals with local translations by using max-pooling layers. The max-pooling operation only retains the maximum feature value within a specified window over a feature map, resulting in shrinkage in the size of a feature map. The shrinking factor is controlled by a hyper-parameters i.e., pooling size, which controls the size of the pooling window and stride for the subsequent window. Let $S \times S$ be the size of a non-pooled feature map and $p$ and $s$ be the pooling size and stride respectively, then max-pooling operation result in a feature map of size $W\times W$, where $W=(S-p)/(s+1)$. The pooling layer computes every point $O_{(i,j)}$ in a feature map $I$, by taking maximum value out of the specific window of length $p$ as;
\begin{equation}
  O_{(i,j)}=max(I_{(i+p,j+p)}) .
  \end{equation}
  
A non-linear activation function is used at the end of the network to convert features in to class probabilities. The softmax activation function converts the output values in to soft class probabilities and is used as non-linearity at the output layer in the proposed architectures. The class with the largest probability after applying softmax is assigned to the central pixel in the corresponding input patch. The probability $P$ of each class $c$ from a number of classes $K$ is given as;
\begin{equation}
 P(y=c|a)= \dfrac{e^{aw_{c}}}{\sum_{k=1}^{K}e^{aw_{k}}} ,
  \end{equation}

where $a$ and $w$ are the feature and weight vectors.

The layers in a CNN can be manipulated in a variety of ways. Similarly, the feature maps produced by a convolution layer can be concatenated with output of another layer in numerous ways. To segment brain tumor, the dependencies between output labels need to be taken into account. Structured approaches such as CRF can be used to model such dependencies. A CRF has been successfully applied to image segmentation task in \citep{55}, but it is computationally expensive, especially when combined with a CNN. This makes it less practical because efficient systems are required, which can work in real time in the medical field. The nexus architectures are formed by combining two CNNs using the concatenating ability of the convolution layer. The output of first network is concatenated with the input of the second network, forming a nexus of two models, hence the name, nexus architecture. The first network takes four input planes of size $33\times33$ from the four MRI modalities and generates an output of size $5\times 15 \times 15$. The output is concatenated with the input to the second network that also takes patches sized $15 \times 15$ from four input MRI modalities. Thus, there are a total of nine input planes at the second module, which produce an output of size $5\times 1 \times 1$. The output represents the probability of the four type of tumors and the normal class. The proposed architecture is computationally efficient and model dependencies between neighbouring pixels. In this work, five type of nexus architectures are proposed namely: linear nexus, two-path nexus, two-path linear nexus, inception nexus and inception linear nexus. The architectures are discussed in detail in the following subsections.

\subsubsection{Linear Nexus (LN)}
\begin{figure*}[t]
  \centering
  \includegraphics[width=5.8in, height=2.25in]{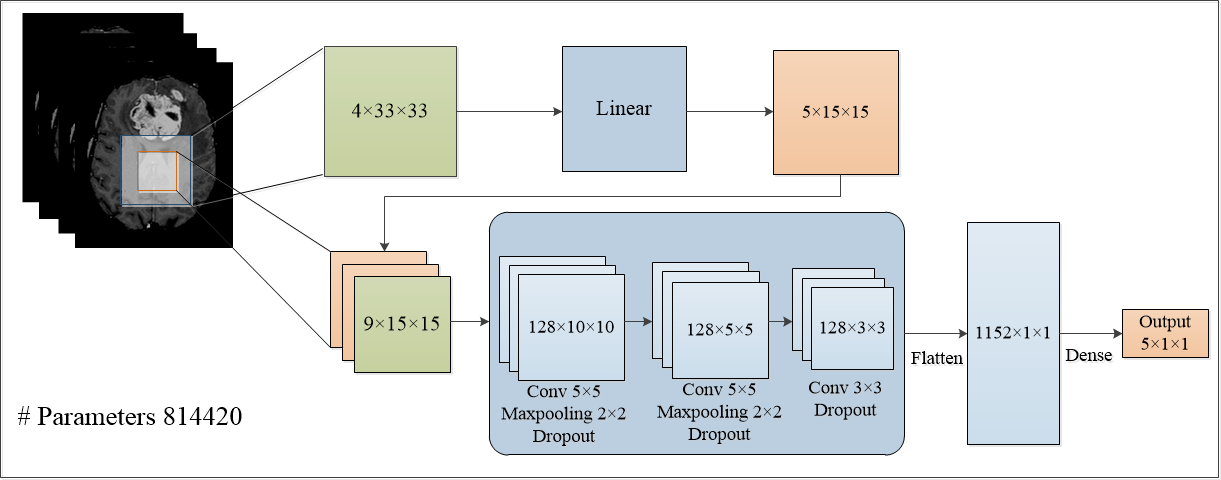}
  \caption{Proposed linear nexus architecture .}\label{fig:fig4}
\end{figure*}

In this architecture, two linear CNNs are concatenated in a cascade manner. The output of the first CNN is simply treated as additional channel to the input of the second CNN, as shown in \autoref{fig:fig4}. The network contains $1152$ features in the fully connected layer. A high value of drop-out is utilized, which reduces the chances of over-fitting as the number of parameters in the network grow. The networks with more features in the fully connected layer tend to be slower, because convolution layers are significantly faster as compared to the fully connected layers. A large number of features in the fully connected layer can lead to better learning. The number of features are selected optimally for the fully connected layer giving sufficiently good results for the segmentation task in a limited time frame.

\subsubsection{Two-path Nexus (TPN)}
\begin{figure*}[t]
  \centering
  \includegraphics[width=5.8in, height=2.4in]{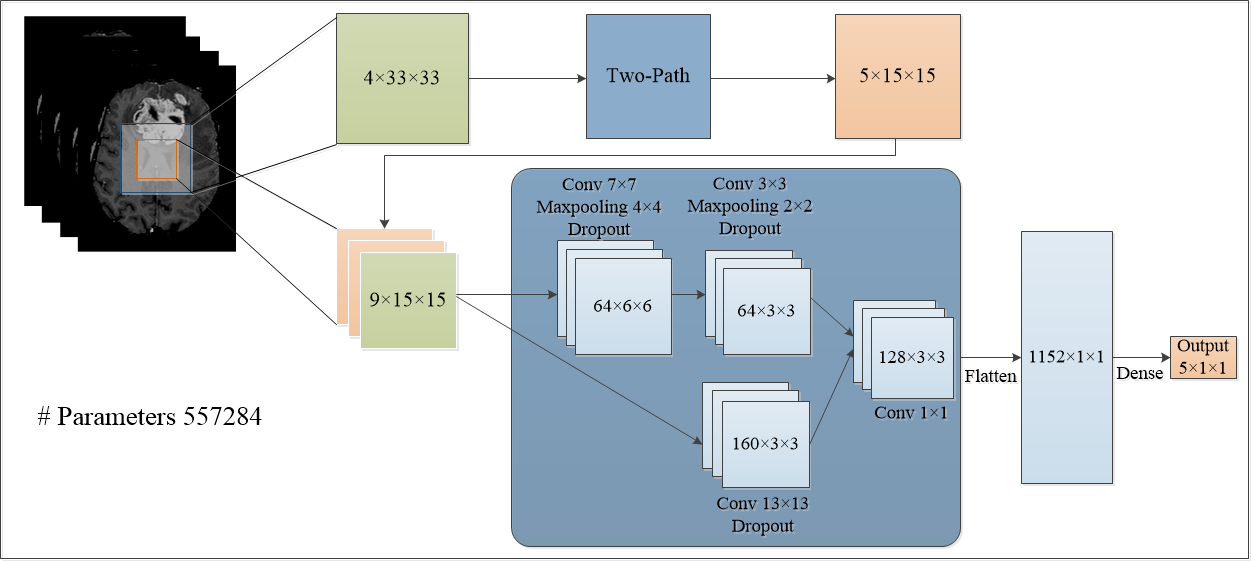}
  \caption{Proposed two-path nexus architecture.}\label{fig:fig5}
\end{figure*}

In this architecture, two-path networks are used, instead of linear CNNs, as shown in \autoref{fig:fig5}. In the first network, input patches are processed in two different paths to get more contextual information from the original data by varying kernel sizes. One of the paths uses kernels of size $13\times13$, while the other path uses smaller kernels of size $7\times7$ and $3\times3$ in two different convolutional layers. Larger kernels are used to extract contextual information, while smaller kernels detect local features. The output of first network is concatenated to the input of the subsequent network. The second network also processes the input in two different streams and combines the results at the end. Similar to the linear nexus architecture, TPN also contains $1152$ features in the fully connected layer. 
\begin{figure*}[t]
  \centering
  \includegraphics[width=5.8in, height=2.1in]{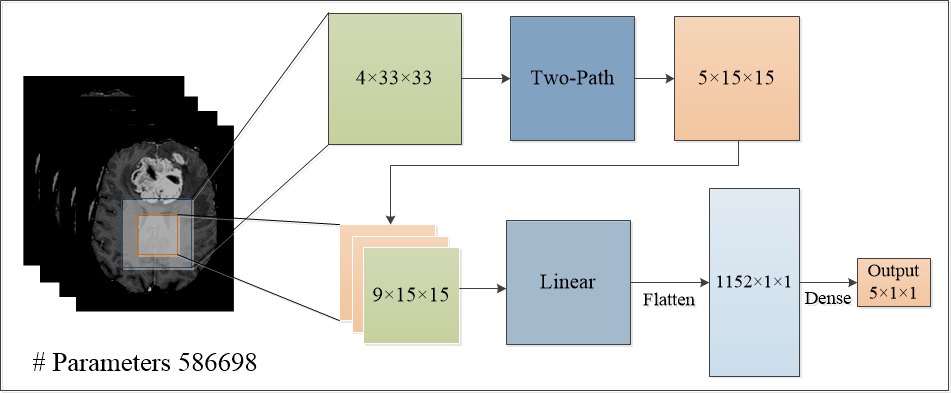}
  \caption{Proposed two-path linear nexus architecture, a combination of two-path and linear nexus.}\label{fig:fig6}
\end{figure*}
\begin{figure*}[!h]
  \centering
  \includegraphics[width=5.8in, height=3.3in]{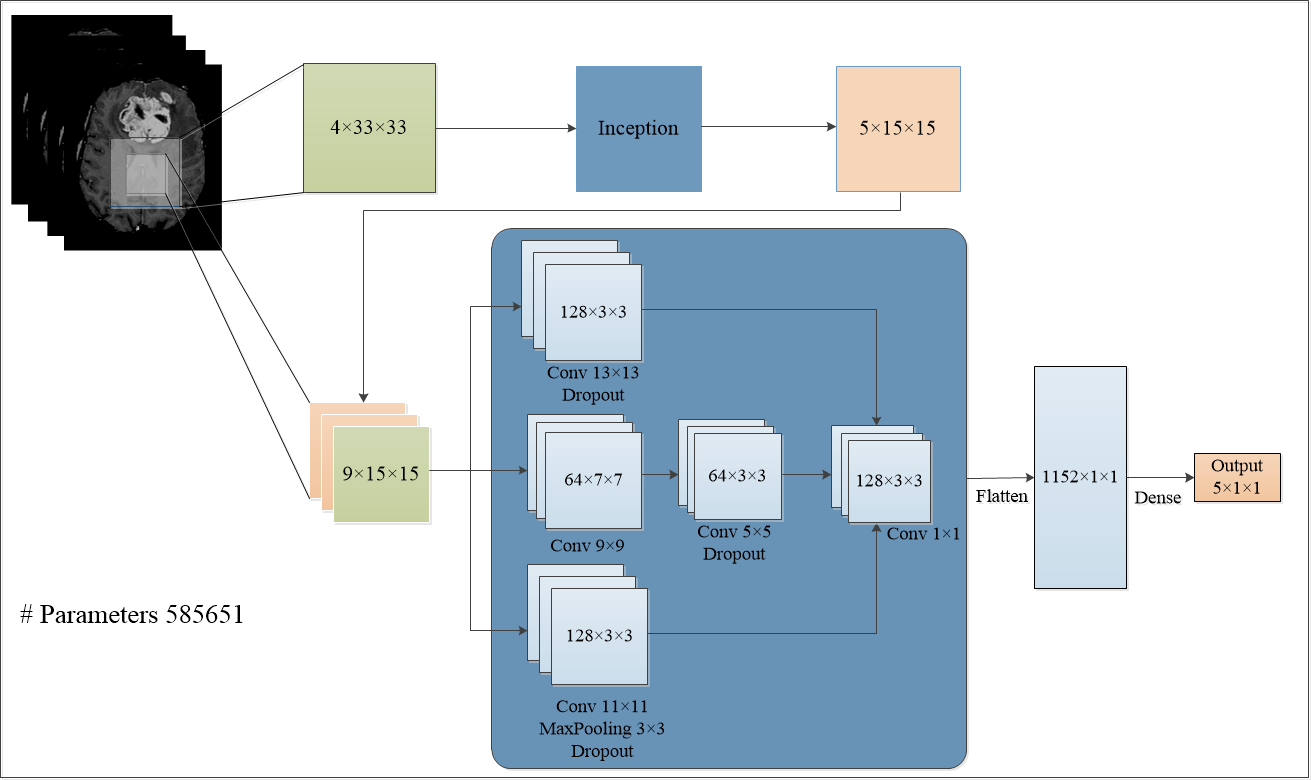}
  \caption{Proposed inception nexus architecture.}\label{fig:fig7}
\end{figure*}

\subsubsection{Two-path Linear Nexus (TLinear)}
The proposed TLinear nexus architecture is a combination of LN and TPN as shown in \autoref{fig:fig6}. This network combines parallel and sequential processing in the initial and later parts of the network, respectively. The first half of TLinear nexus has larger kernels, therefore takes more global information into account, whereas the later part focuses more on local information. This architecture also uses the same number of features in the fully connected layer as in previous two architectures.

\subsubsection{Inception Nexus (IN)}
An inception nexus architecture is formed by combining two inception modules as shown in \autoref{fig:fig7}. The inception module is formed by using three parallel paths, making the network very deep, hence the name inception nexus. The layers in inception module utilize varying kernel sizes in a range of $5\times5$ to $13\times13$, which helps in detecting contextual as well as local information. This architecture is best suited for utilizing parallel processing capability of the available hardware. 
\begin{figure*}[!h]
  \centering
  \includegraphics[width=5.8in, height=2.1in]{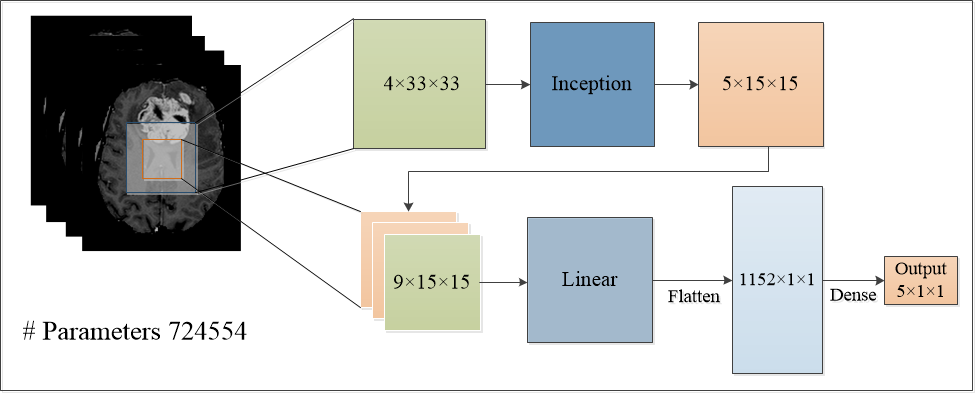}
  \caption{Proposed inception linear nexus architecture, a combination of inception and linear nexus.}\label{fig:fig8}
\end{figure*}

\subsubsection{Inception Linear Nexus (ILinear)}
ILinear architecture is a combination of LN and the IN module, as shown in \autoref{fig:fig8}. Among the proposed architectures, ILinear nexus is the most efficient design, which incorporates both the speed and precision. It has the ability to learn large number of features compared to the TLinear architecture due to increased parallel processing in the first half of nexus. Larger kernels are used in the first part to get more contextual information, on the other hand smaller kernels are utilized in the later part to model dependencies among pixels. The network contains $1152$ features in the fully connected layer.  
 
\subsection{Post-processing}
In the post-processing step, simple morphological operators are used to improve the segmentation results. Due to high intensity around the skull portion, some false positives might appear after segmentation. The opening and closing morphological operators that employ erosion and dilation techniques in succession, are utilized to remove small false positives around the edges of the segmented image.  

\section{Experimental Setup}
\label{sec:experiment}
The details of the dataset used for evaluating the proposed architectures, the parameters and training procedure are discussed in the following subsections. 
\subsection{Dataset}
Experiments are carried out on BRATS $2013$ and BRATS $2015$ datasets \citep{13}, which contain four MRI modalities i.e., T1, T1c, T2 and T2flair, along with segmentation labels for the training data. The BRATS $2013$ dataset contains a total of $30$ training MR images out of which, $20$ belong to HGG and $10$ belong to LGG. The BRATS $2015$ dataset comprises of a total of $274$ training MR images out of which $220$ are HGG and $54$ are LGG images. Pixel labels are only available for the training data and are divided into five classes namely, necrosis, edema, non-enhancing tumor, enhancing tumor and healthy tissues. The proposed methodology is evaluated using three classes i.e., enhancing tumor, core tumor (necrosis, non-enhancing and enhancing tumor) and complete tumor (all tumor classes).

\subsection{Neural Network Parameters}
The proposed methodology is implemented in python using Keras library \citep{52}, which provides numerous methods and pre-trained models to implement the convolutional neural networks over TensorFlow or Theano back-end. It is both GPU and CPU compatible, making it an exceptional tool for deep learning. The grid search algorithm is utilized to tune the hyper-parameters of the network. The kernel weights for the proposed networks are initialized randomly using normal mode initialization \citep{56}. The biases on all layers are set to zero except for the last softmax layer, where it is set to $0.2$. The max-pooling and convolution layers use a stride of $1$, while kernel dimensions and the max-pooling size are shown in \autoref{fig:fig4}-8. Details of different implementation parameters used for the proposed architectures are as follows.

\subsubsection{Neuronal Activation}
A neuronal activation function is used to control the output of neurons in the neural network. Different activation functions such as  max-out, rectified linear units (ReLU), leakyReLU and tangent are analyzed in this study. A  max-out layer \citep{16} takes feature maps produced by a convolution layer as input and selects and retains the map with higher values from adjacent maps. It returns a feature map $FM_{a}$, with maximum feature values compared with all the input maps i.e., ${I_{a},I_{(a+1)},….,I_{(a+k-1)}}$, where 'max' operation is performed over all spatial positions ${(i, j)}$ to get the output feature map as;
\begin{equation}
  FM_{(a,i,j)}=max(I_{(a,i,j)},I_{(a+1,i,j)},….,I_{(a+k-1,i,j)}) . 
  \end{equation}
    
A ReLU activation function has either a zero or a positive value output. It is a non-linear activation function, which calculates a maximum of zero and the input $z$ as follows;
\begin{equation}
  f(z)=max(0,z) .
  \end{equation}
In contrast to ReLU, a LeakyReLU \citep{53} outputs a small value when the input is less than zero and the tangent function outputs values in the interval of $[-1, 1]$. Both leakyReLU and tangent activations resulted in model under-fitting with the validation loss above one. On the other hand,  max-out and ReLU activations have generated validation loss of $0.45$ and $0.4$, respectively. Therefore, ReLU is selected as the activation function for neurons in all of the proposed architectures.
  
\subsubsection{Normalization}
Batch normalization \citep{15} is used to normalize the activations after every batch of input data. Batch normalization utilizes an activation function to maintain the mean and standard deviation of activations near zero and one, respectively. Due to a large learning rate, layer weights change significantly, which can amplify small changes in layer parameters causing the weights to explode. Batch normalization hinders gradient from blowing out of proportions and keeps it within an acceptable limit during back-propagation. A normalized feature map $N$ is obtained as, 
\begin{equation}
 N=nl(BN(W_{a})),
  \end{equation}
where $BN()$ represents the batch normalization on weight parameter $W_{a}$ and $nl$ is the ReLU non-linearity.
 
\subsubsection{Regularizer}
Regularizers are used to reduce over-fitting by imposing penalties on layer weights during network optimization. A Regularizer is incorporated within the loss function and penalizes the parameters such as kernel weights, bias and activity etc. A drop-out layer \citep{14} can be used as a regularizer on feature maps obtained from the convolution layer. This layer drops a certain percentage of activations randomly at each update during training by setting dropped units to zero. This reduces over-fitting as units have to learn independently instead of relying on each other for producing an output. A vector of independent random variables $r_{a}$ is generated using Bernoulli distribution, where each element of vector has a probability $p$ of being equal to $1$. This vector $r_{a}$ is then multiplied with output $O_{a}$ of the preceding layer in an element wise manner to get a sparse output $\tilde{O}_{a}$, which is used as input to next layer in the hierarchy. The element wise multiplication operation is denoted by $\otimes$ and the output of drop-out layer is measured as,  
\begin{equation}
 \tilde{O}_{a}= r_{a} \otimes O_{a}. 
  \end{equation}

   The small number of training patches increase the chances of over-fitting in a convolutional neural network. In this study, a high drop-out value of $0.5$ is used in the first half of the nexus network, whereas a drop-out of $0.4$ is used in the second half. A drop-out value of $0.3$ is used before the last layer of network to drop weak features and produce an output based on strong features.

\subsubsection{Optimizer}
An optimizer is used to compute the loss function at the output layer of the network and distribute the updated values throughout the network. Stochastic gradient descent (SGD) \citep{50} is used along with the loss function in the back-propagation algorithm. The loss function $L_{i}$ is computed as;
\begin{equation}
 L_{i}=\frac{-1}{B} \sum_{i=1}^{B}log(P(Y_{i}=y_{i})) ,
  \end{equation}
where $Y_{i}$ is the target class label, $y_{i}$ is the predicted class label and $B$ represents the mini-batches of data. 

\subsection{Training}
The network training requires maximizing the probability of correct label in the dataset or minimizing loss during training. The CNN outputs the probability distribution over all labels, where the goal is to maximize probability of the true label. SGD algorithm is used for training the network using a weighted two phase training procedure. The methods are discussed in detail in the following subsections.  

\subsubsection{Stochastic Gradient Descent}
The SGD calculates the negative log probability after a batch of data to compute the step size for descent in all weight parameters of the network. Therefore, there is no need to process the whole dataset at a time when mini-batch approach is used, which helps in reducing the memory and processing requirements. This approach is implemented by passing a set of mini-batches through the network at a time and computing the gradient descent for each mini-batch. Gradient change $\Delta W_{i}$ is then propagated back to the entire network, updating weights on all layers and parameters.

The SGD when navigating through the ravines, can get stuck in local optima due to steeper surface curves in certain areas of the descent algorithm. Classical momentum (CM) computes the gradient change after taking a step in the direction of momentum and can amplify the gradient loss, which leads to more erroneous probabilities. To minimize the objective function $f(\theta)$, CM calculates the velocity $V_{t+1}$ for next iteration based on previous velocity $V_{t}$, momentum coefficient $\mu$, learning rate $lr$ and true gradient at current iteration $\nabla f(\theta_{t})$ as, 
\begin{equation}
  V_{t+1}=\mu V_{t}-lr \nabla f(\theta_{t}).
  \end{equation}
  
  
To further optimize the gradient descent algorithm, Nesterov accelerated gradient (NAG) descent algorithm is utilized in this work, which computes the gradient change and then takes a step in the direction of gradient \citep{51}. In this way, the SGD algorithm becomes smarter, as it knows where it is going beforehand compared to the CM, in which the gradient just rolls down the slope. There is a small difference between CM and NAG in the update of velocity vector for the next iteration. Gradient at the current position is tinkered by adding an additional term $\mu V_{t}$ to compute the velocity vector for next iteration. This allows NAG to compute the partial gradient before updating gradient making it more responsive. The velocity vector $V_{t+1}$ and new gradient $\theta_{t+1}$ for NAG are computed as,
  \begin{equation}
  V_{t+1}=\mu V_{t}-lr \nabla f(\theta_{t}+ \mu V_{t}) , 
  \end{equation} 
and, 
\begin{equation}
\theta_{t+1}= \theta_{t}+V_{t+1}  .
  \end{equation}

\subsubsection{Two-Phase Training}
Brain tumor segmentation is a highly data imbalance classification problem, where most of the pixels in brain structure comprise of healthy tissues, which makes it difficult to train the network as training with equivalent classes increases prediction bias towards the tumor classes. Whereas, training with the true distribution overwhelms the network with healthy patches causing it to be biased towards the healthy pixels. To deal with the data imbalance problem, a two-phase training process is presented. In the first phase, labels are given an equal representation in the training data, while the data are distributed unevenly in the second phase. The network is trained using the equal distribution for $20$ epochs on two hundred thousand training examples. In the second phase, labels are represented by their true distribution where $98\%$ of the labels in an image belong to healthy pixels and the remaining $2\%$ are divided among the tumor classes. 

A weighted training technique is used for second phase, where data belonging to different classes are assigned weights according to their distribution in the training data. For instance, class $0$ i.e., normal class is assigned a weight of $8$, class $2$ i.e., edema is assigned a weight of $2$ and classes $1$, $3$ and $4$ i.e., core tumor are assigned a weight of $1$. In this way, healthy patches impact on training weights is nearly eight times more than the tumor patches. Therefore, the effect of true distribution is maintained, while using equal number of patches from individual classes. In the second phase, the output layer of the network is trained for $5$ epochs on thirty thousand training patches, while keeping all the other layers fixed. Only the output layer is tuned to the true distribution of labels, where weights of all the other layers are not trained. In this way, most of the capacity of network is used equally for all classes, whereas the output layer is tuned for actual frequency of labels in the data making it more useful for problem at hand.

As the concatenating utility of convolution layer is used, it is necessary to match the output dimensions of first CNN with the input dimensions of the second CNN. To accomplish this, a larger input i.e., $33\times33$ is passed to the first CNN resulting in an output of $15\times15$, which is the input dimension for the second CNN. The momentum coefficient $\mu=0.9$ is utilized in this work along with a variable learning rate. The initial learning rate is set to $lr=0.01$, which is gradually decreased to $lr=0.01 \times 10^{-4}$ for the optimal performance.

\subsection{Evaluation Parameters}
Segmentation results are evaluated based on three metrics namely; dice similarity coefficient (DSC), sensitivity, and specificity \citep{13}. All metrics are computed for three classes i.e., complete tumor, core tumor and the enhancing tumor. Dice score is computed by overlapping predicted output image $L$ with the manually segmented label $G$. The pixel by pixel intersection of two images result in the dice similarity coefficient and is given as,
\begin{equation}
DSC = 2 \times \dfrac{|G \cap L|}{|G|+|L|} . 
  \end{equation}
Sensitivity is a measure of accuracy of correctly classified tumor labels. It determines, how well the model has performed to detect tumor in a given image and is given as,
\begin{equation}
Sensitivity = \dfrac{|L| \cap |G|}{|G|}   .
  \end{equation}
  Specificity computes the accuracy of correctly classified labels of normal output class. It determines, how well the model has performed to specify tumor to its labelled area. It is computed by taking intersection of predicted normal label $P_{o}$ with actual normal label $L_{o}$ as,
  \begin{equation}
Specificity = \dfrac{|L_{o} | \cap |G_{o} |}{|G_{o}|}   . 
  \end{equation}
  
\section{Experimental Results and Discussion}
\label{sec:results}
The results obtained on BRATS $2013$ and BRATS $2015$ datasets for the proposed architectures are reported in this section. As BRATS $2013$ dataset has varying resolution in the third dimension, 2D slices are used for training and testing purposes. All the results reported in this study are based on ReLU used as activation function in all the convolution layers. Drop-out regularization and batch normalization are performed on the output feature maps of convolution layers. All the models used in this study, contain a drop-out and batch normalization layer after every convolution layer. It is observed that by only using a drop-out layer before the output layer leads to over-fitting to training and validation data, which is a misleading effect as it gives good validation accuracy but poor testing accuracy. This effect is significantly evident in BRATS $2015$ dataset, as it contains a large number of images compared to the BRATS $2013$ dataset. Hence, training on lower number of patches causes the model to be incapable of learning enough features to segment all images accurately. The proposed model learns enough features and is generalized by using drop-out after each convolution layer giving good accuracy on the validation set. 

The data is divided into training and testing sets randomly, where training data consist of $80$ percent of the images in both datasets. The testing set consists of $20$ percent of the data and is used for validating the proposed methodology. The proposed architectures are trained and tested on a Dell core-i7 machine with a RAM of $16$ GB.  
\begin{figure}[!t]
  \centering
  \includegraphics[width=3.2in, height=1.6in]{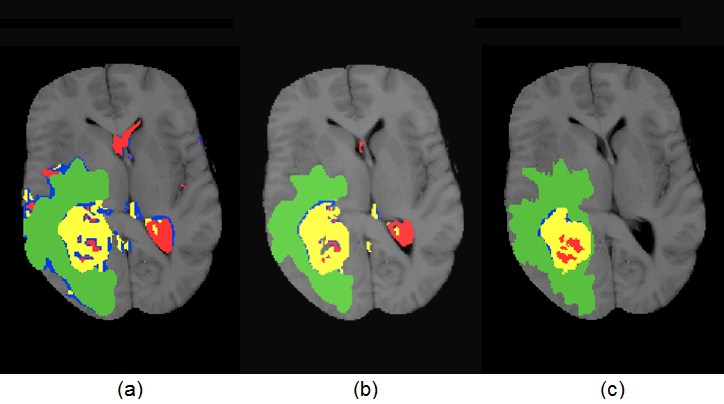}
  \caption{Segmentation results on MICCAI BRATS $2013$ dataset using (a) one phase training, (b) two phase training, while (c) is the ground truth.}\label{fig:fig9}
\end{figure}

\begin{table*}[b]
\caption{Segmentation results in terms of dice score, sensitivity and specificity on MICCAI BRATS $2013$ dataset using one phase training method.}
\label{table:tab1}
\centering
\scalebox{0.85}
{
  \begin{tabular}{c | c | c c c | c c c | c c c}
    \hline
 Training   & \multirow{2}{*}{Models} &
      \multicolumn{3}{c}{Dice} &
      \multicolumn{3}{c}{Sensitivity} &
      \multicolumn{3}{c}{Specificity} \\ \cline{3-11}
   & & Complete & Core & Enhancing & Complete & Core & Enhancing & Complete & Core & Enhancing  \\
    \hline \hline
  & LN &	0.73 &	0.70 &	0.74 &	0.72 &	0.68 &	0.75 &	0.87 &	0.82 &	0.81 \\ 
    & TPN &	0.69 &	0.75 &	0.78 & 	0.69 &	0.76 & 	0.80 &	0.84 &	0.80 &	0.81 \\ 
single-phase    & TLinear &	0.73 &	0.77 &	0.81 &	0.75 &	0.80 &	0.83 &	0.86 &	0.82 &	0.83 \\ 
    & IN &	0.68 &	0.78 &	0.80 &	0.70 &	0.76 &	0.83 &	0.84 &	0.79 &	0.79 \\ 
   & ILinear &	0.75 &	0.78 &	0.82 &	0.78 &	0.82 &	0.85 &	0.87 &	0.81 &	0.84 \\ \hline \hline
   
   & LN &	0.85 &	0.85 &	0.84 &	0.86 &	0.84 &	0.86 &	0.93 &	0.91 &	0.91 \\ 
    & TPN &	0.83 &	0.87 &	0.89 & 	0.84 &	0.87 & 	0.92 &	0.91 &	0.91 &	0.89 \\ 
 two-phase    & TLinear &	0.86 &	0.87 &	0.91 &	0.86 &	0.89 &	0.94 &	0.92 &	0.91 &	0.90 \\ 
    & IN &	0.81 &	0.85 &	0.93 &	0.82 &	0.88 &	0.95 &	0.86 &	0.89 &	0.88 \\ 
    & ILinear &	0.87 &	0.89 &	0.92 &	0.90 &	0.89 &	0.95 &	0.94 &	0.93 &	0.92 \\ \hline

  \end{tabular}
}
\end{table*}

\subsection{Effect of using Two-Phase Training}
The models are trained using a two-phase training method discussed in Section $3.3.2$. The improvements resulting from the training methodology are shown in \autoref{fig:fig9}, where without the second phase of training, the number of false positives increase and labels are distributed more evenly. In comparison, when models are trained using the two-phase training procedure, it removes most of the false positives and the labels are also distributed according to their overall distribution. This makes models more capable of handling data imbalance problem. 

The experimental results for the proposed architectures when applied on BRATS $2013$ data using the first and second phase of training are shown in \autoref{table:tab1}. It is evident from these results that two-phase training results in improved performance parameters. The lack of false positives are indicated by high specificity values, which are low, when only single phase training is utilized. The proposed model predicts the tumor labels according to the data distribution, solidifying its utility in medical image analysis. It is evident from these results that models trained using a single phase of training under perform in various performance parameters, which is not acceptable in the clinical environment.
   
\begin{figure*}[!b]
  \centering
  \includegraphics[width=5in, height=4in]{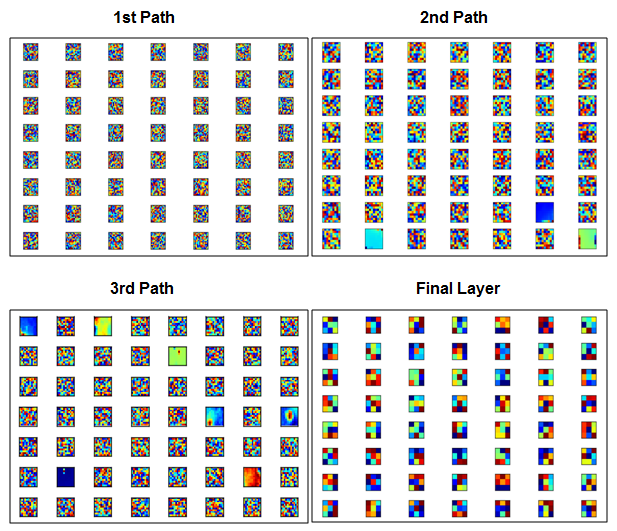}
  \caption{Randomly selected kernels from first layer of the three paths and the final convolution layer in the ILinear Nexus. The initial kernels extract localized information, whereas, the final layer kernels are edge detectors.}\label{fig:fig10}
\end{figure*}
\subsection{Multipath Nexus}
It is observed that by increasing the paths in a network, the learning capability of the network increases. The proposed architectures use parallel processing thus reducing time overhead and improving accuracy of results. It is observed that no matter how many layers are stacked linearly, results improve minimally after a certain threshold, which is found to be four layers in our case. Therefore, the proposed architectures are formed using parallel paths and the results depict the improvement in accuracy of segmentation. The five architectures i.e. linear, TLinear, two-path, ILinear and inception nexus, each has an increased amount of parallel processing, respectively. The results also stop improving after three parallel layers similar to the case of increasing linearity after a certain threshold. Hence, the ILinear nexus with three parallel layers achieves the optimal performance. 

\begin{figure*}[!t]
  \centering
  \includegraphics[width=6in, height=3in]{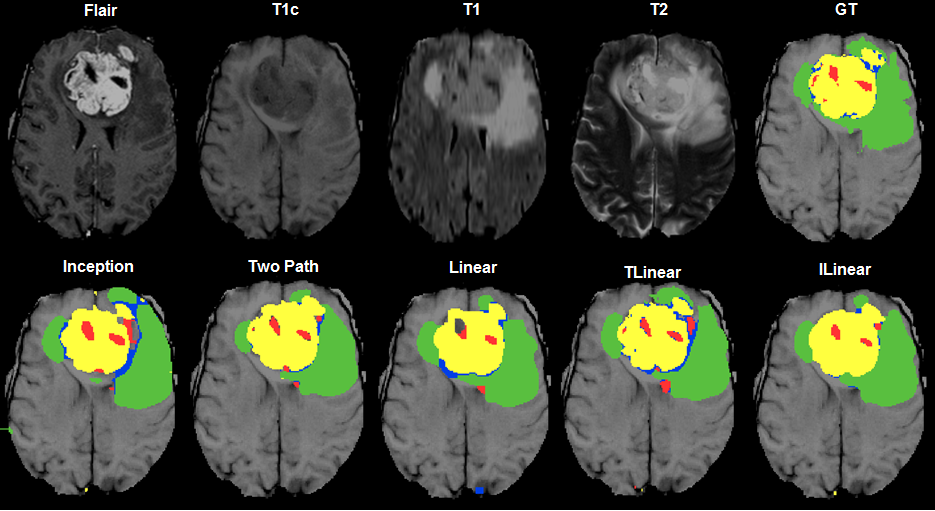}
  \caption{Segmentation results on MR images from MICCAI BRATS $2013$ dataset. Top row (left to right): four input modalities (Flair, T1c, T1, T2) and ground truth (GT). Bottom row (left to right): segmentation results from proposed architectures i.e. Inception, two-path, linear, TLinear and ILinear nexus respectively. The red, green, blue and yellow colors represent necrosis (label $1$), edema (label $2$), non-enhancing (label $3$) and enhancing (label $4$) tumors respectively.}
\label{fig:fig11}
\end{figure*}
Randomly selected kernels from the first three layers of the ILinear nexus architecture are shown in \autoref{fig:fig10}. The first three images refer to the first convolution layer in each path in the first half of the ILinear nexus, whereas the final image represents the final convolution layer in the second part of the network. Large filters learn more of the local features, whereas the smaller kernels focus more on edge detection. The first path has a filter size of $13\times13$, therefore, it has more distributed kernels. On the other hand, the final layer has a kernel size of $3\times3$, therefore, it has more edge detectors. The proposed architectures give excellent results in core and enhancing regions of tumor as well as in specifying tumor to its true location. The high specificity values indicate the lack of false positives, therefore, these models have proven to be good in predicting true negatives accurately.

The inception nexus architecture gives improved performance on core and enhancing metrics, whereas for the complete tumor, the performance is slightly lower. On the other hand, the linear model works well for the complete tumor metric, but the results are slightly lower in other metrics. Therefore, the hybrid model i.e., ILinear is proposed and stands out among all architectures in all performance metrics. The inception nexus faces difficulty in learning effective features of edema labelled as $2$, which is the major tumor class. Although, this effect is not as prominent in BRATS $2013$ dataset as it is in BRATS $2015$ dataset, but it still persists in the architectures, mainly due to the lower number of training examples in BRATS $2013$ dataset. The segmentation results of the proposed architectures are shown in \autoref{fig:fig11}, highlighting the effectiveness of combing parallel and linear CNNs via TLinear and ILinear nexus architectures. The top row shows the four MRI modalities used and the ground truth, whereas the bottom row shows the segmentation results for the proposed architectures.

\begin{table*}[!t]
\caption{A comparison of the proposed architectures with state-of-the-art techniques in terms of dice score, sensitivity and specificity for MICCAI BRATS $2013$ dataset using two-phase training method.}
\label{table:tab2}
\centering
\scalebox{0.88}
{
  \begin{tabular}{c| c c c |c c c | c c c}
    \hline
    \multirow{2}{*}{Models} &
      \multicolumn{3}{c}{Dice} &
      \multicolumn{3}{c}{Sensitivity} &
      \multicolumn{3}{c}{Specificity} \\ \cline{2-10}
    & Complete & Core & Enhancing & Complete & Core & Enhancing & Complete & Core & Enhancing  \\
    \hline \hline
    Cordier &	0.84 &	0.68 &	0.65 &	0.88 &	0.63 &	0.68 &	0.81 &	0.82 &	0.66 \\ 
   Doyle &	0.71 &	0.46 &	0.52 &	0.66 &	0.38 &	0.58 &	0.87 &	0.70 &	0.55 \\ 
    Festa &	0.72 &	0.66 &	0.67 &	0.77 &	0.77 &	0.70 &	0.72 &	0.60 &	0.70 \\ 
    Meier &	0.82 &	0.73 &	0.69 &	0.76 &	0.78 &	0.71 &	0.92 &	0.72 &	0.73 \\ 
    Reza &	0.83 &	0.72 &	0.72 &	0.82 &	0.81 &	0.70 &	0.86 &	0.69 &	0.76 \\ 
    Zhao (II) &	0.84 &	0.70 &	0.65 &	0.80 &	0.67 &	0.65 &	0.89 &	0.79 &	0.70 \\ 
    Tustison &	0.87 &	0.78 &	0.74 &	0.85 &	0.74 &	0.69 &	0.89 &	0.88 &	0.83 \\ 
    Proposed LN &	0.85 &	0.85 &	0.84 &	0.86 &	0.84 &	0.86 &	0.93 &	0.91 &	0.91 \\ 
    Proposed TPN &	0.83 &	0.87 &	0.89 & 	0.84 &	0.87 & 	0.92 &	0.91 &	0.91 &	0.89 \\ 
    Proposed TLinear &	0.86 &	0.87 &	0.91 &	0.86 &	0.89 &	0.94 &	0.92 &	0.91 &	0.90 \\ 
    Proposed IN &	0.81 &	0.85 &	0.93 &	0.82 &	0.88 &	0.95 &	0.86 &	0.89 &	0.88 \\ 
   \rowcolor{lightgray} Proposed ILinear &	0.87 &	0.89 &	0.92 &	0.90 &	0.89 &	0.95 &	0.94 &	0.93 &	0.92 \\ \hline
  \end{tabular}
}  
\end{table*}

\subsection{Results Comparison}
\begin{figure*}[!t]
  \centering
\includegraphics[width=2in, height=1.9in]{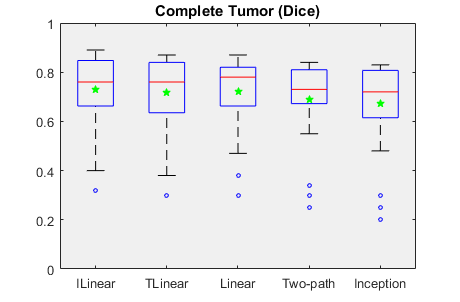} \includegraphics[width=2in, height=1.9in]{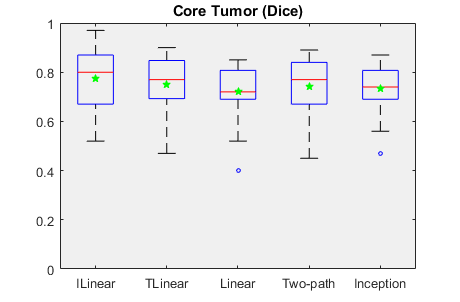} \includegraphics[width=2in, height=1.9in]{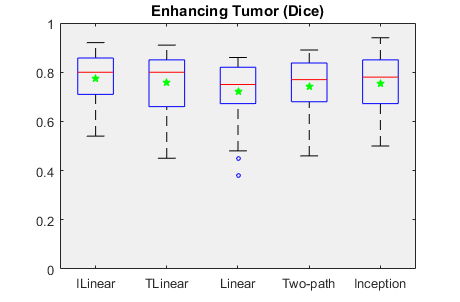}
\includegraphics[width=2in, height=1.9in]{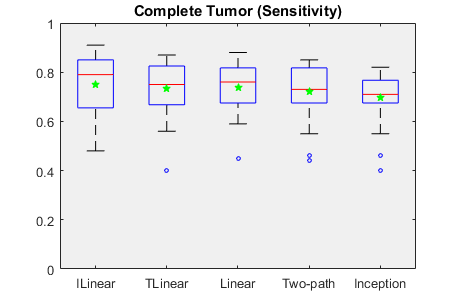} \includegraphics[width=2in, height=1.9in]{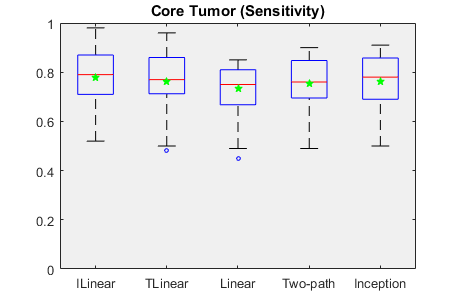} \includegraphics[width=2in, height=1.9in]{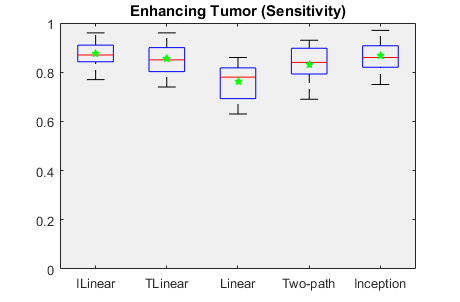}
\includegraphics[width=2in, height=1.9in]{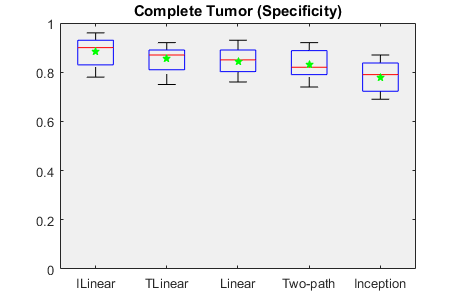} \includegraphics[width=2in, height=1.9in]{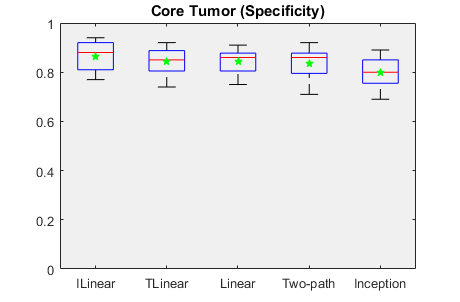} \includegraphics[width=2in, height=1.9in]{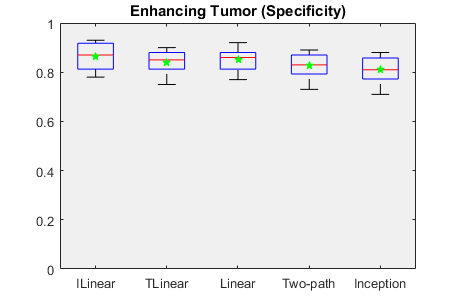}  
  \caption{Boxplot of the results on MICCAI BRATS $2013$ dataset using proposed architectures: ILinear, TLinear, linear, two-path and inception nexus, in terms of dice score, sensitivity and specificity. The arithematic mean is presented in green, the median in red and the outliers in blue.}\label{fig:fig12}
\end{figure*}
The results of this study for BRATS $2013$ dataset, are compared with the state-of-the-art techniques that have been reported in BRATS challenge benchmark \citep{13} and are presented in \autoref{table:tab2}. The proposed architectures outperform state-of-the-art techniques in terms of core and enhancing tumor regions, while producing comparable results for the whole tumor region. The improved performance on core and enhancing classes is the highlight of this work along with the high specificity values. As label $2$ represents the majority in tumor classes, it affects the whole tumor score more than other labels. Since, label $2$ is not incorporated in the core and enhancing tumor region, the proposed models have achieved excellent results for these regions. It is also important to note that the best architecture reported in \citep{13}, takes up to $100$ minutes to segment the brain tumor using a CPU cluster. In comparison, the proposed algorithm takes $5-10$ minutes to segment the whole brain by using CPU. Due to the significance of brain tumor segmentation in medical industry, the time constraint is of immense importance along with the accuracy and precision. These results show the effectiveness and computational efficiency of the proposed architectures.

The overall performance of the proposed architectures i.e., ILinear, TLinear, linear, two-path and inception over the entire dataset is summarized in \autoref{fig:fig12}. It is evident that apart from some outliers, the models perform well over all the images in the dataset, especially in specifying the tumor to its actual region and in the categories of core and enhancing tumor metrics. In comparison to state-of-the-art-techniques that have focused more on whole tumor region, the proposed method not only gives better results in the whole tumor region, but it also outperforms methods in literature by a large margin in core and enhancing tumor regions.

\subsection{Results on BRATS 2015 Dataset}
\begin{table*}[!t]
\caption{Segmentation results of the proposed architectures in terms of dice score, sensitivity and specificity on MICCAI BRATS $2015$  dataset using the two phase training method.}
\label{table:tab3}
\centering
\scalebox{1}
{
  \begin{tabular}{c| c c c |c c c |c c c}
    \hline
    \multirow{2}{*}{Models} &
      \multicolumn{3}{c}{Dice} &
      \multicolumn{3}{c}{Sensitivity} &
      \multicolumn{3}{c}{Specificity} \\ \cline{2-10}
    & Complete & Core & Enhancing & Complete & Core & Enhancing & Complete & Core & Enhancing  \\
    \hline \hline
    LN &	0.84 &	0.82 &	0.83 &	0.80 &	0.83 &	0.83 &	0.92 &	0.91 &	0.89 \\ 
    TPN &	0.82 &	0.83 &	0.89 &	0.79 &	0.84 &	0.89 &	0.91 &	0.90 &	0.92 \\ 
    TLinear &	0.83 &	0.87 &	0.89 &	0.85 &	0.87 &	0.92 &	0.92 &	0.91 &	0.90 \\ 
    IN &	0.79 &	0.86 &	0.90 &	0.78 &	0.88 &	0.91 &	0.88 &	0.89 &	0.91 \\ 
   \rowcolor{lightgray} ILinear &	0.86 &	0.87 &	0.90 &	0.86 &	0.86 &	0.94 &	0.93 &	0.93 &	0.92 \\ \hline
  \end{tabular}
}
\end{table*}
\begin{figure*}[!t]
  \centering
  \includegraphics[width=6in, height=3in]{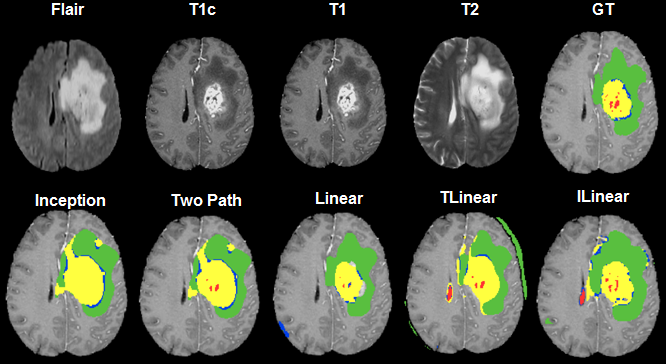}
  \caption{Segmentation results on MR images from MICCAI BRATS $2015$ dataset. Top row (left to right): four input modalities (Flair, T1c, T1, T2) and ground truth (GT). Bottom row (left to right): segmentation results from proposed architectures i.e. Inception, two-path, linear, TLinear and ILinear nexus respectively. The red, green, blue and yellow colors represent necrosis (label $1$), edema (label $2$), non-enhancing (label $3$) and enhancing (label $4$) tumors respectively.}\label{fig:fig13}
\end{figure*}
The experimental results on BRATS $2015$ data are obtained using two phase training procedure and are reported in \autoref{table:tab3}. The single phase training on this dataset does not yield any significant results. The linearity in architecture allows it to learn better features for class labelled $2$, whereas non-linearity allows the architecture to learn better features for all the other classes, leading to the better core and enhancing scores. The proposed architectures are particularly good in specifying tumor to its substantial area generating high specificity values. The ILinear nexus architecture gives the best performance on BRATS $2015$ dataset due to its ability to learn features of all classes effectively. A visual illustration of the segmentation results for BRATS $2015$ dataset is shown in \autoref{fig:fig13} for a random image from the test set. The top row shows the four MRI modalities used and the ground truth, whereas the second row shows the segmentation results for the proposed architectures.

\begin{table*}[!t]
\caption{A comparison of the proposed architectures with state-of-the-art techniques in terms of dice score for MICCAI BRATS $2015$ dataset.}
\label{table:tab4}
\centering
\scalebox{1}
{
  \begin{tabular}{c| c c c}
    \hline
    \multirow{2}{*}{Models} & \multicolumn{3}{c}{Dice} \\ \cline{2-4} 
    & Complete & Core & Enhancing \\ \hline \hline
    DCNN (Tseng Kuan Lun) &	0.92 &	0.85 &	0.72 \\ 
    CLDF (Loic Le Folgoc) &	0.79 &	0.67 &	0.70 \\ 
    VCTB (Edgar A. Rios Piedra) &	0.74 &	0.54 &	0.54 \\ 
    AG (Bi Song) &	0.85 &  	0.70 &	0.73 \\ 
    3DNet\_2 (Adria Casamitjana) &	0.89 &	0.76 &	0.37 \\ 
    FCNN (Xiaomei Zhao) &	0.80 &	0.68 &	0.65  \\ 
    Proposed LN &	0.84 &	0.82 &	0.83  \\ 
    Proposed TPN &	0.82 &	0.83 &	0.89 \\ 
    Proposed TLinear &	0.83 &	0.87 &	0.89 \\ 
    Proposed IN &	0.79 &	0.86 &	0.90 \\ 
   \rowcolor{lightgray} Proposed ILinear &	0.86 &	0.87 &	0.90 \\ \hline
  \end{tabular}
}
\end{table*}

It is worth mentioning that to the best of our knowledge, no significant results have been reported on BRATS $2015$ dataset. Most of the results report the DICE score \citep{54}, which is used for comparison as shown in \autoref{table:tab4}. A deep convolutional neural network has been proposed by Tseng Kuan Lun that focuses on global information in contrast to the patch based approach and has achieved the best results for complete tumor region. Loic Le Folgoc has proposed an improved implementation of the decision forests, called a cascade of lifted decision forests (CLDF). A technique arising from variable nature of brain images and variability characterization of tumor boundaries (VCTB) has been presented by Edgar A. Rios Piedra. Bi Song has used anatomy guided (AG) technique to find the regions of interest and has then utilized random forests to compute final segmentation. A 3D CNN (3DNet\_2) has been proposed by Adria Casamitjana, which has achieved the second best results in \citep{54} on BRATS $2015$ dataset. Three different 3D architectures have been proposed by Adria Casamitjana, but only the best is used for comparison in this work. Xiaomei Zhao has proposed the use of CRFs alongside a fully convolution neural network to segment brain tumor regions. It is evident from the results that most of the models and techniques focus on whole or complete tumor dice score, whereas enhancing tumor region suffers from the worst results. In contrast, the proposed methodology outperforms state-of-the-art techniques in core and enhancing regions while giving comparable results for the complete tumor region.

\section[CONCLUSION]{Conclusion}
\label{sec:conclusion}
Automated methods for brain tumor segmentation can play a vital role in future diagnostic procedures for the brain tumor. In this paper, an automated method for segmenting brain tumor using deep convolutional neural networks has been presented. Different architectural settings have been explored and the effect of using multiple parallel paths in an architecture are validated. The segmentation results on two datasets i.e., BRATS $2013$ and BRATS $2015$ verifies that the proposed architecture improves the performance when compared with state-of-the-art techniques. Two networks are stacked over one another to form a new ILinear nexus architecture, which achieves the best results among the proposed architectures. The first network in ILinear nexus architecture contains parallel placement of layers, whereas in the second network, layers are arranged linearly. A two-phase weighted training procedure helps the network to adapt to the class imbalance problem in the data. The proposed network performs satisfactorily on both the datasets for whole tumor score, while producing outstanding results for core and enhancing regions. The proposed architecture has also been validated to be exceptional in predicting true negatives and specifying tumor to its substantial area, generating high specificity values. The capacity analysis of neural networks suggest that the performance could improve further, provided an increased number of training examples and computational resources.
\bibliographystyle{unsrt}
\bibliography{refr}

\begin{thebibliography}{10}

\bibitem{1}
Stefan Bauer, Roland Wiest, Lutz-P Nolte, and Mauricio Reyes.
\newblock A survey of mri-based medical image analysis for brain tumor studies.
\newblock {\em Physics in medicine and biology}, 58(13):R97, 2013.

\bibitem{2}
American Brain~Tumor Assocation.
\newblock {\em About Brain Tumors: A Primer for Patients and Caregivers}.
\newblock American Brain Tumor Association, 2013.

\bibitem{3}
David~N Louis, Hiroko Ohgaki, Otmar~D Wiestler, Webster~K Cavenee, Peter~C
  Burger, Anne Jouvet, Bernd~W Scheithauer, and Paul Kleihues.
\newblock The 2007 who classification of tumours of the central nervous system.
\newblock {\em Acta neuropathologica}, 114(2):97--109, 2007.

\bibitem{4}
Maria-del-Mar Inda, Rudy Bonavia, Joan Seoane, et~al.
\newblock Glioblastoma multiforme: a look inside its heterogeneous nature.
\newblock {\em Cancers}, 6(1):226--239, 2014.

\bibitem{5}
Dietmar Krex, Barbara Klink, Christian Hartmann, Andreas von Deimling, Torsten
  Pietsch, Matthias Simon, Michael Sabel, Joachim~P Steinbach, Oliver Heese,
  Guido Reifenberger, et~al.
\newblock Long-term survival with glioblastoma multiforme.
\newblock {\em Brain}, 130(10):2596--2606, 2007.

\bibitem{6}
Mohammad Havaei, Axel Davy, David Warde-Farley, Antoine Biard, Aaron Courville,
  Yoshua Bengio, Chris Pal, Pierre-Marc Jodoin, and Hugo Larochelle.
\newblock Brain tumor segmentation with deep neural networks.
\newblock {\em Medical image analysis}, 35:18--31, 2017.

\bibitem{7}
Ghazaleh Tabatabai, Roger Stupp, Martin~J van~den Bent, Monika~E Hegi,
  J{\"o}rg~C Tonn, Wolfgang Wick, and Michael Weller.
\newblock Molecular diagnostics of gliomas: the clinical perspective.
\newblock {\em Acta neuropathologica}, 120(5):585--592, 2010.

\bibitem{8}
Lisa~M DeAngelis.
\newblock Brain tumors.
\newblock {\em New England Journal of Medicine}, 344(2):114--123, 2001.

\bibitem{9}
Solmaz Abbasi and Farshad~Tajeri Pour.
\newblock A hybrid approach for detection of brain tumor in mri images.
\newblock In {\em Biomedical Engineering (ICBME), 2014 21th Iranian Conference
  on}, pages 269--274. IEEE, 2014.

\bibitem{10}
Saurabh~A Shah and Narendra~C Chauhan.
\newblock An automated approach for segmentation of brain mr images using
  gaussian mixture model based hidden markov random field with expectation
  maximization.
\newblock {\em Journal of Biomedical Engineering and Medical Imaging}, 2(4):57,
  2015.

\bibitem{11}
Nelly Gordillo, Eduard Montseny, and Pilar Sobrevilla.
\newblock State of the art survey on mri brain tumor segmentation.
\newblock {\em Magnetic resonance imaging}, 31(8):1426--1438, 2013.

\bibitem{12}
T~Logeswari.
\newblock Automatic brain tumor detection through mri--a survey.
\newblock {\em Digital Image Processing}, 8(9):303--305, 2016.

\bibitem{13}
Bjoern~H Menze, Andras Jakab, Stefan Bauer, Jayashree Kalpathy-Cramer, Keyvan
  Farahani, Justin Kirby, Yuliya Burren, Nicole Porz, Johannes Slotboom, Roland
  Wiest, et~al.
\newblock The multimodal brain tumor image segmentation benchmark (brats).
\newblock {\em IEEE transactions on medical imaging}, 34(10):1993--2024, 2015.

\bibitem{17}
Paul Aljabar, R~Heckemann, Alexander Hammers, Joseph Hajnal, and Daniel
  Rueckert.
\newblock Classifier selection strategies for label fusion using large atlas
  databases.
\newblock {\em Medical Image Computing and Computer-Assisted
  Intervention--MICCAI 2007}, pages 523--531, 2007.

\bibitem{19}
Bjoern Menze, Andras Jakab, Stefan Bauer, Mauricio Reyes, Marcel Prastawa, and
  Koen Van~Leemput.
\newblock Proceedings of the miccai challenge on multimodal brain tumor image
  segmentation (brats) 2012, 2012.

\bibitem{18}
Adriano Pinto, S{\'e}rgio Pereira, Higino Correia, Jorge Oliveira, Deolinda~MLD
  Rasteiro, and Carlos~A Silva.
\newblock Brain tumour segmentation based on extremely randomized forest with
  high-level features.
\newblock In {\em Engineering in Medicine and Biology Society (EMBC), 2015 37th
  Annual International Conference of the IEEE}, pages 3037--3040. IEEE, 2015.

\bibitem{20}
Kwokleung Chan, Te-Won Lee, Pamela~A Sample, Michael~H Goldbaum, Robert~N
  Weinreb, and Terrence~J Sejnowski.
\newblock Comparison of machine learning and traditional classifiers in
  glaucoma diagnosis.
\newblock {\em IEEE Transactions on Biomedical Engineering}, 49(9):963--974,
  2002.

\bibitem{21}
Marcel Prastawa, Elizabeth Bullitt, Sean Ho, and Guido Gerig.
\newblock A brain tumor segmentation framework based on outlier detection.
\newblock {\em Medical image analysis}, 8(3):275--283, 2004.

\bibitem{22}
Mohammad Havaei, Pierre-Marc Jodoin, and Hugo Larochelle.
\newblock Efficient interactive brain tumor segmentation as within-brain knn
  classification.
\newblock In {\em Pattern Recognition (ICPR), 2014 22nd International
  Conference on}, pages 556--561. IEEE, 2014.

\bibitem{23}
Andac Hamamci, Nadir Kucuk, Kutlay Karaman, Kayihan Engin, and Gozde Unal.
\newblock Tumor-cut: segmentation of brain tumors on contrast enhanced mr
  images for radiosurgery applications.
\newblock {\em IEEE transactions on medical imaging}, 31(3):790--804, 2012.

\bibitem{24}
Jens Kleesiek, Armin Biller, Gregor Urban, U~Kothe, Martin Bendszus, and
  F~Hamprecht.
\newblock Ilastik for multi-modal brain tumor segmentation.
\newblock {\em Proceedings of MICCAI 2013 Challenge on Multimodal Brain Tumor
  Segmentation (BRATS 2013)}, 2014.

\bibitem{25}
Raphael Meier, Stefan Bauer, Johannes Slotboom, Roland Wiest, and Mauricio
  Reyes.
\newblock Appearance-and context-sensitive features for brain tumor
  segmentation.
\newblock {\em Proceedings of MICCAI BRATS Challenge}, pages 020--026, 2014.

\bibitem{26}
Nicholas~J Tustison, KL~Shrinidhi, Max Wintermark, Christopher~R Durst,
  Benjamin~M Kandel, James~C Gee, Murray~C Grossman, and Brian~B Avants.
\newblock Optimal symmetric multimodal templates and concatenated random
  forests for supervised brain tumor segmentation (simplified) with antsr.
\newblock {\em Neuroinformatics}, 13(2):209--225, 2015.

\bibitem{27}
Stefan Bauer, Lutz-P Nolte, and Mauricio Reyes.
\newblock Fully automatic segmentation of brain tumor images using support
  vector machine classification in combination with hierarchical conditional
  random field regularization.
\newblock In {\em International Conference on Medical Image Computing and
  Computer-Assisted Intervention}, pages 354--361. Springer, 2011.

\bibitem{29}
Karan Sikka, Nitesh Sinha, Pankaj~K Singh, and Amit~K Mishra.
\newblock A fully automated algorithm under modified fcm framework for improved
  brain mr image segmentation.
\newblock {\em Magnetic Resonance Imaging}, 27(7):994--1004, 2009.

\bibitem{30}
Darko Zikic, Ben Glocker, Ender Konukoglu, Antonio Criminisi, C~Demiralp, Jamie
  Shotton, O~Thomas, Tilak Das, Raj Jena, and S~Price.
\newblock Decision forests for tissue-specific segmentation of high-grade
  gliomas in multi-channel mr.
\newblock {\em Medical Image Computing and Computer-Assisted
  Intervention--MICCAI 2012}, pages 369--376, 2012.

\bibitem{36}
Yoshua Bengio, Aaron Courville, and Pascal Vincent.
\newblock Representation learning: A review and new perspectives.
\newblock {\em IEEE transactions on pattern analysis and machine intelligence},
  35(8):1798--1828, 2013.

\bibitem{33}
Liang-Chieh Chen, George Papandreou, Iasonas Kokkinos, Kevin Murphy, and Alan~L
  Yuille.
\newblock Semantic image segmentation with deep convolutional nets and fully
  connected crfs.
\newblock {\em arXiv preprint arXiv:1412.7062}, 2014.

\bibitem{34}
Evan Shelhamer, Jonathan Long, and Trevor Darrell.
\newblock Fully convolutional networks for semantic segmentation.
\newblock {\em IEEE transactions on pattern analysis and machine intelligence},
  39(4):640--651, 2017.

\bibitem{Valverde2017159}
Sergi Valverde, Mariano Cabezas, Eloy Roura, Sandra González-Villà, Deborah
  Pareto, Joan~C. Vilanova, Lluís Ramió-Torrentà, Àlex Rovira, Arnau
  Oliver, and Xavier Lladó.
\newblock Improving automated multiple sclerosis lesion segmentation with a
  cascaded 3d convolutional neural network approach.
\newblock {\em NeuroImage}, 155:159 -- 168, 2017.

\bibitem{Pan201788}
Xipeng Pan, Lingqiao Li, Huihua Yang, Zhenbing Liu, Jinxin Yang, Lingling Zhao,
  and Yongxian Fan.
\newblock Accurate segmentation of nuclei in pathological images via sparse
  reconstruction and deep convolutional networks.
\newblock {\em Neurocomputing}, 229:88 -- 99, 2017.
\newblock Advances in computing techniques for big medical image data.

\bibitem{Qayyum2017}
Adnan Qayyum, Syed~Muhammad Anwar, Muhammad Awais, and Muhammad Majid.
\newblock Medical image retrieval using deep convolutional neural network.
\newblock {\em Neurocomputing}, pages~--, 2017.

\bibitem{35}
Le~Hou, Dimitris Samaras, Tahsin~M Kurc, Yi~Gao, James~E Davis, and Joel~H
  Saltz.
\newblock Patch-based convolutional neural network for whole slide tissue image
  classification.
\newblock In {\em Proceedings of the IEEE Conference on Computer Vision and
  Pattern Recognition}, pages 2424--2433, 2016.

\bibitem{37}
Yuhong Li, Fucang Jia, and Jing Qin.
\newblock Brain tumor segmentation from multimodal magnetic resonance images
  via sparse representation.
\newblock {\em Artificial intelligence in medicine}, 73:1--13, 2016.

\bibitem{38}
Mark Lyksborg, Oula Puonti, Mikael Agn, and Rasmus Larsen.
\newblock An ensemble of 2d convolutional neural networks for tumor
  segmentation.
\newblock In {\em Scandinavian Conference on Image Analysis}, pages 201--211.
  Springer, 2015.

\bibitem{39}
Jun Jiang, Yao Wu, Meiyan Huang, Wei Yang, Wufan Chen, and Qianjin Feng.
\newblock 3d brain tumor segmentation in multimodal mr images based on learning
  population-and patient-specific feature sets.
\newblock {\em Computerized Medical Imaging and Graphics}, 37(7):512--521,
  2013.

\bibitem{40}
V~Rao, M~Shari~Sarabi, and A~Jaiswal.
\newblock Brain tumor segmentation with deep learning.
\newblock {\em MICCAI Multimodal Brain Tumor Segmentation Challenge (BraTS)},
  pages 56--59, 2015.

\bibitem{41}
Pavel Dvorak and Bjoern Menze.
\newblock Structured prediction with convolutional neural networks for
  multimodal brain tumor segmentation.
\newblock {\em Proceeding of the Multimodal Brain Tumor Image Segmentation
  Challenge}, pages 13--24, 2015.

\bibitem{14}
Nitish Srivastava, Geoffrey~E Hinton, Alex Krizhevsky, Ilya Sutskever, and
  Ruslan Salakhutdinov.
\newblock Dropout: a simple way to prevent neural networks from overfitting.
\newblock {\em Journal of Machine Learning Research}, 15(1):1929--1958, 2014.

\bibitem{15}
Sergey Ioffe and Christian Szegedy.
\newblock Batch normalization: Accelerating deep network training by reducing
  internal covariate shift.
\newblock {\em arXiv preprint arXiv:1502.03167}, 2015.

\bibitem{16}
Ian~J Goodfellow, David Warde-Farley, Mehdi Mirza, Aaron Courville, and Yoshua
  Bengio.
\newblock Maxout networks.
\newblock {\em arXiv preprint arXiv:1302.4389}, 2013.

\bibitem{42}
G~Collewet, M~Strzelecki, and F~Mariette.
\newblock Influence of mri acquisition protocols and image intensity
  normalization methods on texture classification.
\newblock {\em Magnetic Resonance Imaging}, 22(1):81--91, 2004.

\bibitem{43}
Nicholas~J Tustison, Brian~B Avants, Philip~A Cook, Yuanjie Zheng, Alexander
  Egan, Paul~A Yushkevich, and James~C Gee.
\newblock N4itk: improved n3 bias correction.
\newblock {\em IEEE transactions on medical imaging}, 29(6):1310--1320, 2010.

\bibitem{44}
Steve Pieper, Bill Lorensen, Will Schroeder, and Ron Kikinis.
\newblock The na-mic kit: Itk, vtk, pipelines, grids and 3d slicer as an open
  platform for the medical image computing community.
\newblock In {\em Biomedical Imaging: Nano to Macro, 2006. 3rd IEEE
  International Symposium on}, pages 698--701. IEEE, 2006.

\bibitem{45}
Andriy Fedorov, Reinhard Beichel, Jayashree Kalpathy-Cramer, Julien Finet,
  Jean-Christophe Fillion-Robin, Sonia Pujol, Christian Bauer, Dominique
  Jennings, Fiona Fennessy, Milan Sonka, et~al.
\newblock 3d slicer as an image computing platform for the quantitative imaging
  network.
\newblock {\em Magnetic resonance imaging}, 30(9):1323--1341, 2012.

\bibitem{48}
Geoffrey~W Burr, Robert~M Shelby, Severin Sidler, Carmelo Di~Nolfo, Junwoo
  Jang, Irem Boybat, Rohit~S Shenoy, Pritish Narayanan, Kumar Virwani,
  Emanuele~U Giacometti, et~al.
\newblock Experimental demonstration and tolerancing of a large-scale neural
  network (165 000 synapses) using phase-change memory as the synaptic weight
  element.
\newblock {\em IEEE Transactions on Electron Devices}, 62(11):3498--3507, 2015.

\bibitem{55}
Konstantinos Kamnitsas, Christian Ledig, Virginia~FJ Newcombe, Joanna~P
  Simpson, Andrew~D Kane, David~K Menon, Daniel Rueckert, and Ben Glocker.
\newblock Efficient multi-scale 3d cnn with fully connected crf for accurate
  brain lesion segmentation.
\newblock {\em Medical Image Analysis}, 36:61--78, 2017.

\bibitem{52}
P.W.D. Charles.
\newblock Project title.
\newblock \url{https://github.com/charlespwd/project-title}, 2013.

\bibitem{56}
Akira Kasahara.
\newblock Nonlinear normal mode initialization and the bounded derivative
  method.
\newblock {\em Reviews of Geophysics}, 20(3):385--397, 1982.

\bibitem{53}
S{\'e}rgio Pereira, Adriano Pinto, Victor Alves, and Carlos~A Silva.
\newblock Brain tumor segmentation using convolutional neural networks in mri
  images.
\newblock {\em IEEE transactions on medical imaging}, 35(5):1240--1251, 2016.

\bibitem{50}
Yoshua Bengio.
\newblock Practical recommendations for gradient-based training of deep
  architectures.
\newblock In {\em Neural networks: Tricks of the trade}, pages 437--478.
  Springer, 2012.

\bibitem{51}
Ilya Sutskever, James Martens, George~E Dahl, and Geoffrey~E Hinton.
\newblock On the importance of initialization and momentum in deep learning.
\newblock {\em ICML (3)}, 28:1139--1147, 2013.

\bibitem{54}
K~Farahani BH~Menze, M~Reyes and D~Kwon J~Kalpathy-Cramer.
\newblock Proceedings of miccai-brats 2016, 2016.

\end{thebibliography}
\end{document}